%% file: arxiv.tex
\documentclass{article}

\usepackage{booktabs}
\usepackage{graphicx}
\usepackage{graphicx}
\usepackage{caption} 
\usepackage{xcolor}
\usepackage[preprint]{neurips_2023}
\usepackage{wrapfig}

\usepackage[utf8]{inputenc} 
\usepackage[T1]{fontenc}    
\usepackage{hyperref}       
\usepackage{url}            
\usepackage{booktabs}       
\usepackage{amsfonts}       
\usepackage{nicefrac}       
\usepackage{microtype}      
\usepackage{xcolor}         
\usepackage{algorithm}
\usepackage{algpseudocode}

\usepackage{amsmath, amssymb, amsthm, mathtools} 
\usepackage{wrapfig}  
\usepackage{mdwlist}  
\usepackage{sidecap}  
\usepackage{bbm}
\usepackage{bm,amsbsy} 

\usepackage{subcaption}  
\usepackage{caption}
\usepackage{cancel} 
\usepackage{ulem} 

\usepackage{stmaryrd}
\usepackage{MnSymbol}
\input{notations}

\title{Pre-Pruning and Gradient-Dropping Improve Differentially Private Image Classification}

\author{%
  Kamil Adamczewski \\
  ETH Zurich, D-ITET\\
  MPI for Intelligent Systems, EI \\
  \texttt{kamil.m.adamczewski@gmail.com} \\
  \And
  Yingchen He \\
 Computer Science Dep. \\
  Univ. British Columbia \\
  \And
  Mijung Park \\
Computer Science Dep. \\
  Univ. British Columbia \\
  \texttt{mijungp@cs.ubc.ca} \\
}

\begin{document}

\maketitle

\begin{abstract}
%
%

Scalability is a significant challenge when it comes to applying differential privacy to training deep neural networks. The commonly used DP-SGD algorithm struggles to maintain a high level of privacy protection while achieving high accuracy on even moderately sized models. 
To tackle this challenge, we take advantage of the fact that neural networks are overparameterized, which allows us to improve neural network training with differential privacy.
Specifically, we introduce a new training paradigm that uses \textit{pre-pruning} and \textit{gradient-dropping} to reduce the parameter space and improve scalability. The process starts with pre-pruning the parameters of the original network to obtain a smaller model that is then trained with DP-SGD. During training, less important gradients are dropped, and only selected gradients are updated. Our training paradigm introduces a tension between the rates of pre-pruning and gradient-dropping, privacy loss, and classification accuracy. Too much pre-pruning and gradient-dropping reduces the model's capacity and worsens accuracy, while training a smaller model requires less privacy budget for achieving good accuracy.  
We evaluate the interplay between these factors and demonstrate the effectiveness of our training paradigm for both training from scratch and fine-tuning pre-trained networks on several benchmark image classification datasets. The tools can also be readily incorporated into existing training paradigms. 
%

\end{abstract}

\section{Introduction}

Differential privacy (DP) \citep{Dwork14} is a gold standard privacy notion that is widely used in many applications in machine learning.
Generally speaking, the \textit{accuracy} of a model trained with a dataset trades off with the guaranteed level of \textit{privacy} of the individuals in the dataset. 
One of the most popular differentially private (DP) training algorithms for deep learning is \textit{DP-SGD} \citep{DP_SGD}, which adds a carefully calibrated amount of noise to the gradients during training using stochastic gradient descent (SGD) to provide a certain level of differential privacy guarantee for the trained model at the end of the training. Since it adds noise to the gradients in every training step, its natural consequence is that the accuracy of the model drops.  
Typically, the trade-off between privacy and accuracy gets worse  when the model size gets larger, as the dimension of the gradients increases accordingly.

In the current state of differentially private classifier training, the performance between small and large models with and without differential privacy guarantees is striking. A small two-layer convolutional neural network trained on MNIST data with $\epsilon=2$ provides a classification accuracy of $95\%$ \citep{Da21}. However, a larger and more complex architecture, the ResNet-50 model, containing $50$-layers, trained on ImageNet data with $\epsilon < 5$ provides a mere classification accuracy of $2.4\%$, while the same model without privacy guarantee achieves $75\%$ accuracy \citep{jax_dp}. 
Given that this is more or less the current state of the art in the literature (without relying on public data), we cannot help but notice a disappointingly considerable gap between the differentially privately trained and non-privately trained classifiers.

Why is there such a large gap between what the  current state-of-the-art deep learning can do and what the  current state-of-the-art differentially private deep learning can do? The reason is that those deep models that can reliably classify high-dimensional images are \textit{large-scale} and DP-SGD does not scale well. The definition of DP requires that every parameter dimension has to be equally guarded. 
For example, consider two models, where the first model has $10$ parameters and the second model has $1000$ parameters, where these parameters are normalized by some constant (say $1$) to have a limited sensitivity, which is required for the DP guarantee. As we need to add an equal amount of independent noise to each of the  parameters to guarantee DP, where the parameters are normalized by the same constant, the signal-to-noise ratio per parameter dimension becomes roughly $100$ times worse in the second model with $1000$ parameters compared to the first model with $10$ parameters. 

Existing work attempts to reduce this gap by utilizing access to public data which is similar in some sense to their private data.
These roughly fall into two categories. In the first category, 
the training paradigm proceeds with pre-training the large classifier with public data and then fine-tuning the whole network or some selected portion of the network with DP-SGD using private data. 
In the second category, the training paradigm uses the public data to reduce the dimension of the parameters to improve the aforementioned signal-to-noise ratio in DP-SGD training.  See \secref{related_work} for the descriptions of the papers under these categories. 

In this work, we also pursue the idea of reducing the parameter dimension. However, we use neural network pruning techniques to study the impact of reducing the parameter space on differentially private training. In particular, we apply pruning in two forms: (1) \textbf{pre-pruning} where the set of trainable network parameters is decreased, prior to the private training, and (2) \textbf{gradient-dropping}, where the parameter space is preserved but only a subset of weights is updated at every iteration. Both forms can be combined into one paradigm which can be easily implemented with existing training paradigms, and can be used both when fine-tuning pre-trained networks and training from scratch. 

Experimental results 
prove to narrow the gap between the performance of privately trained models and  that of non-privately trained models by improving the privacy-accuracy trade-offs in training large-scale classifiers.
In what comes next, we start by describing relevant background information before introducing our method. 


\section{Background}
This work builds on the ideas from neural network pruning. While neural network pruning has a broad range of research outcomes, we summarize a few relevant works that help understand the core ideas of our training paradigm. We then also briefly describe differential privacy for an unfamiliar audience.  

\subsection{Neural Network Pruning}
\label{subsec:pruning}
The main reasoning behind neural network pruning is that neural networks are typically considered over-parametrized, and numerous observations indicate that pruning out non-essential parameters does not sacrifice  a model's performance by much. There are two types of pruning techniques: one-shot pruning and iterative pruning. 

In one-shot pruning, a model is given, and based on some criteria, the parameters of the model are pruned out \textit{all at once}. This one-shot pruning can be done in a \textit{post-hoc} or \textit{foresight} manner. In the \textit{post-hoc} one-shot pruning, we first train a model and then discard less important parameters. In the \textit{foresight} one-shot pruning, before training starts, the parameters of the base model at initialization (the model one plans to train) are pruned out based on some criteria, which yields a sub-architecture of the original model. This one-shot pruning is the idea we will use in the pre-pruning step. In contrast to one-shot pruning, iterative pruning gradually removes a fraction of the parameters in multiple iterations. This idea loosely relates to gradient-dropping which we will describe in \secref{gradient-dropping}.

The main difference between pruning literature and our work is that we do not intend to compress the model and therefore do not remove weights. Instead, we aim to reduce the gradient space during training to improve the training in a differentially private manner and lessen the impact of the privatization mechanism on the final accuracy.




\subsection{Differential privacy}

A mechanism $\mathcal{M}$, that is a modification applied to a model, provides a privacy guarantee if the output of the mechanism given a dataset $\Dat$ is \textit{similar} to that given a neighboring dataset $\Dat'$ where neighboring means $\Dat$ and $\Dat'$ have one sample difference. The similarity is defined by a selected value of $\epsilon$ and $\delta$, which quantify the privacy guarantees. 
Mathematically, when the following equation holds, we say $\mathcal{M}$ is  ($\epsilon$, $\delta$)-DP: 
$
\Pr[\mathcal{M}(\Dat) \in S] \leq e^\epsilon \cdot \Pr[\mathcal{M}(\Dat') \in S] + \delta,
$
 where the inequality holds for all possible sets of the mechanism's outputs $S$ and all neighboring datasets $\Dat$, $\Dat'$.
%

The \textit{Gaussian mechanism} is one of the most widely used DP mechanisms. As the name suggests, the Gaussian mechanism adds noise drawn from a Gaussian distribution, where the noise scale is proportional to two quantities, the first one is, so-called, {\it{global sensitivity}} (which we denote by $\Delta$) and the second one is a privacy parameter (which we denote by $\sigma$). 
The global sensitivity of a function $g$
\citep{dwork2006our} denoted by $\Delta_g$ is defined by the maximum difference in terms of $L_2$-norm $||g(\Dat)-g(\Dat') ||_2$, for neighbouring $\Dat$ and $\Dat'$. 
In the next section, we will describe the relationship between global sensitivity and gradient clipping in the case of neural networks.
The privacy parameter $\sigma$ is a function of $\epsilon$ and $\delta$. Given a selected value of $\epsilon$ and $\delta$, one can readily compute $\sigma$ using numerical methods, e.g., 
the auto-dp package by \citet{wang2019subsampled} or the jax privacy package by \citet{jax-privacy2022github}.

It is worth noting two  properties of differential privacy.  The \textit{post-processing invariance} property \cite{dwork2006our} states that the composition of any data-independent mapping with an $(\epsilon,\delta)$-DP algorithm is also $(\epsilon,\delta)$-DP. 
%
%
The
\textit{composability} property \cite{dwork2006our} states that the strength of the privacy guarantee degrades in a measurable way with repeated use of the training data. 
%

\subsection{Differentially private stochastic gradient descent (DP-SGD)}


Suppose a neural network model with an arbitrary architecture denoted by $f$, and parameterized by $\theta$. In the DP-SGD algorithm \citep{DP_SGD}, 
the gradient of the loss $l$  is computed. However, to ensure privacy, we need to bind the effect of each individual data point's participation in the gradient computation (i.e., to achieve a limited global sensitivity). For this, we bound the $L2$-norm of the sample-wise gradients by some pre-chosen value, the so-called, clip-norm $C$. As such, the sample-wise gradient of a data sample $b$ is computed, $\nabla^C[b] = \frac{\partial}{\partial \theta} l(b)$ where $\nabla^C[b]$ is norm-clipped such that $\|\frac{\partial}{\partial \theta} l(b) \|_2 \leq C$ for all $b$ in batch $\mathcal{B}$ to ensure the limited sensitivity of the gradients. We then add noise to the average gradient using the aforementioned Gaussian mechanism such that
\begin{align}
    \theta^{(t+1)} \; & \leftmapsto \;  \theta^{(t)} - \frac{\eta }{B} \left[ \sum_{b \in \mathcal{B}} \nabla^C[b] + \sigma C \zeta \right]
\end{align} where $\eta$ is the learning rate, the number of samples in a batch $|\mathcal{B}|=B$, and $\sigma$ is the privacy parameter, which is a function of a desired DP level $(\epsilon, \delta)$ (larger value of $\sigma$ indicates a higher level of privacy), and $\zeta$ is the noise such that $\zeta \sim \Nrm(0, I)$. Note that higher clipping norms $C$ result in higher levels of noise added to the gradient.  

Due to the composability of differential privacy, the cumulative privacy loss increases as we increase the number of training epochs (i.e., as we access the training data more often with a higher number of epochs). When models are large-scale, the required number of epochs until convergence increases as well, which imposes more challenges in achieving a good level of accuracy at a reasonable level of privacy. Our method, which we introduce next,  provides a solution to tackle this challenge.

\section{Method}

We propose a differentially private training framework that first (a)  reduces the size of a large-scale classification model by means of pre-pruning and then (b) trains the parameters of the resulting smaller model by updating only a few selected gradients at a time using DP-SGD. The two-step pipeline above is general for any way of pre-pruning and gradient dropping methods. 

Notation-wise, for a given layer, let $I$ be a list of tensors such that a tensor $i \in I$ and $i \in \mathbb{R}^4$ for convolutional layers where the indices of $i$ describe output, input, and two-dimensional kernel dimensions, and $i \in \mathbb{R}^2$ for fully-connected layers, where the indices describe output and input dimensions. Below we describe the procedure in detail and provide exact solutions to how sparsifying the training may work the best.

\subsection{Pre-pruning}
\label{sec:pre-pruning}

The first tool we use to reduce the parameter space is pre-pruning before training starts. Pre-pruning (sometimes referred in the literature as foresight pruning) stands for reducing the model size before the training begins. The idea stems from the so-called lottery hypothesis \citep{frankle2018lottery} which assumes there are subnetworks within the original network model which can be trained from scratch and obtain similar performance as the original large model. 

We look at the process of pre-pruning from the differential privacy perspective, which requires privatizing every training step where data is accessed. We select three diverse methods which are particularly effective and suitable for the differentially private pre-pruning, SNIP \cite{Lee_et_al_2019}, Synflow \cite{Synflow} and random pre-pruning baseline. SNIP requires privatization which we describe below, while the latter two do not utilize data for pre-pruning, which makes it particularly advantageous for differentially private training. 
Pre-pruning selects a set of parameters which are discarded based on a given criterion. Let the indices of the discarded parameters be $I_{pp} \in I$.  We provide the details of how each of the methods selects the parameters to be removed and create a slimmer network below.

\paragraph{Random Pre-pruning} In random pre-pruning, 
given an initial model (or a pre-trained model) and a pruning ratio $p \in [0,1)$, a fraction of $p$ parameters for each layer is drawn uniformly at random and removed. This pruning does not require looking at the data at all, and therefore does not incur any privacy loss at this stage. However, the remaining parameters are not necessarily more informative than those removed. 

\paragraph{Pre-pruning with Synflow}
\citet{Synflow} presents a foresight pruning method, called \textit{Synflow}, based on the concept of synaptic flow, which is a measure of the flow of information through the connections in a neural network. 
The core idea of Synflow is to iteratively minimize the divergence between the synaptic flow in the original network and the pruned network. The algorithm calculates a saliency score for each connection based on the conservation of synaptic flow and prunes the connections with the lowest saliency values, \textit{without looking at any data}. Mathematically, the synaptic saliency score is:
\begin{equation}\label{eq:synaptic_saliency}
S_{SF} = \frac{\partial R_{SF}}{\partial \vtheta} \odot \vtheta.
\end{equation}
Here, $R_{SF}$ is the loss function defined as:
$
R_{SF} = \vone^T (\prod_{l=1}^L \|\vtheta^{l}\|)\vone$
where $\vone$ is the all ones vector and $\|\vtheta^{l}\|$ is the element-wise absolute value of the parameters in the $l^{th}$ layer. 
Synflow is extremely useful for differential private training, as this method is data-free, incurring zero privacy loss  in pre-pruning. 

\paragraph{Pre-pruning with SNIP}
SNIP \citep{Lee_et_al_2019} explores connection sensitivity, which provides an estimate of how the loss function $\LL$ would change if a particular connection is removed from the network.
Formally, the effect of removing a connection $j$ can be measured by:
    $\Delta \LL_j(\vtheta;\Dat) = \LL(\vone \odot \vtheta; \Dat) - \LL((\vone-\ve_j) \odot \vtheta; \Dat)$
where $\vone$ is a vector of ones, $\ve_j$ is the indicator vector for element $j$ (all zeros except the $j$th entry being $1$), $\vtheta$ is a vector of parameters, $\odot$ is an element-wise multiplication, and $\Dat$ represents data.
Then the connection sensitivity is computed as:
$s_j = \frac{\|g_j(\vtheta;\Dat)\|}{\sum_{k=1}^m \|g_k(\vtheta;\Dat)\|}$
where $g_j(\vtheta;\Dat)$ is the derivative of $\mathcal{L}$ with respect to $c_j$. 

Note that SNIP pruning requires data for computing the connection sensitivity (albeit it requires only one epoch of training). We apply DP-SGD on $g_j$, given as $\tilde{g}_j = \frac{1}{B}clip_C(g_j) + \frac{\sigma C}{B}\xi$, where $\xi \sim \Nrm(0, I)$. The complete algorithm for differentially-private SNIP is given in detail in the Appendix.  



\subsection{Gradient-dropping}
\label{sec:gradient-dropping}

Once the network size is reduced, we apply the second tool, gradient-dropping similar to \citep{Grad_Drop}\footnote{While we are updating only selected gradients and zero-ing out unselected gradients, \citet{Grad_Drop} carry residuals which are added to the next gradient, before dropping again.}. Gradient-dropping selects which parameters are updated at a given training step while dropping the others from the update. 

As a result of this selection step, we split the index set $I$ of all the weights into two groups. The first group is an index set $I_{s}$ that contains all the \textit{selected} indices for the gradients that we will update in the following step; another index set $I_{gd}$ that contains all the \textit{dropped} indices for the gradients that we will  not update. Moreover, define a dropping rate $p$ as the fraction of the gradients which are discarded. Then $I = I_{s} \cup I_{gd}$

At a training time $t$, we subsample a mini-batch $\mathcal{B}^{(t)}$ from the private dataset and compute the gradients on the private mini-batch.
As in DP-SGD, we clip the sample-wise gradients. However, for each $b \in \mathcal{B}$, we clip \textit{only} the selected weights indicated by the index set $I_s$:
\begin{align}
    \nabla^{C} &= \frac{\nabla^{C}[I_s, b]}{ \max\left(1, \frac{|\nabla^{C}[I_s, b]|}{C}\right)}.
\end{align} 
The gradients by the index set $I_{gd}$ are zeroed out. 

\paragraph{Gradient-dropping with random and magnitude-based selection}

Unlike pruning which is done only once, gradient dropping is performed at every gradient update. We distinguish two criteria that are needed for effective differentially-private gradient -dropping. 1) It should require a minimum amount of computation not to hamper the training time. 2) It should not require access to the gradients to reduce the impact on the privacy budget. Two methods that satisfy these criteria are random selection and selection based on the parameter magnitude. 

Random dropping removes a fixed portion of parameters to be updated, while other gradients are dropped in a non-DP setting. This criterion has an advantage in that it does not require looking at the data for the selection, i.e., no privacy loss incurred due to the selection. 

In neural network compression \textit{magnitude-based selection} is a common metric for reduing size of a network. Parameters that have high magnitudes are typically considered to be more important for changing the model's output in non-DP settings. We also test this method in the DP-setting when it comes to gradient-dropping. Since in each training step we privatize the gradients, the magnitude of the parameters is also DP, with respect to the training data, due to the post-processing invariance.

At each gradient update step, we select the indices which correspond to the weights with small magnitude. Then, as these weights are considered unimportant we do not update them and therefore we remove the gradients that correspond to those weights. Note that unlike in the magnitude pruning, we do not remove those weights from the model. 


\subsection{Combined DP-SSGD}

Having described the two tools in the previous sections, we combine them to form the \textit{DP-SSGD (differentially-private sparse stochastic gradient descent)} algorithm. 

Pre-pruning and gradient-dropping may be applied together to reinforce the specification of the DP training and decrease the rate of redundancies in the model during the training. The two tools are used subsequently in the process. First, the network is pruned with a given pre-pruning procedure and the pre-pruning rate $\pi_{pi} \in [0,1]$. The indices of the pre-pruned weights, $I_{pp}$ are permanently stored-away resulting in a model with a smaller set of effective parameters. Secondly, the remaining model is trained with DP-SGD, where gradient-dropping is applied. At every iteration, given pruning rate $\pi_{gd} \in [0,1]$, a different set of indices $I_{gd}$ is selected. Then the indices described by $I_{pp}$ and $I_{gd}$ are combined as the gradients removed $I_{rm}$ and only the remaining gradients $ I \setminus I_{rm}$ are applied at every step

\subsection{Discussion on the benefits of reducing the gradient space in the DP-training}

As a result of removing gradients at every step, the set of non-zero values decreases, and thereby the norm of the gradient set, which is of paramount importance for the effective differentially-private training. Smaller gradient norm benefits the DP-training in two crucial ways. Firstly, notice that decreasing the size of the gradients will be clipped only if $|\nabla^{C}[I_s, b]| > C$. Moreover, they are clipped proportionally to the value $\frac{|\nabla^{C}[I_s, b]|}{C}$, therefore a small gradient norm results in less clipping. Reducing the parameter size results in decreasing the norm and preserving larger gradient values. In other words, we trade off small gradients for keeping the large signal. 

Whether we use the full-size gradients or the gradients of selected gradients, if we use the same clipping norm $C$, the sensitivity of the sum of the gradients in both cases is simply $C$.  
However, adding noise to a vector of longer length (when we do not select weights) has a worse signal-to-noise ratio than adding noise to a vector of shorter length (when we do select weights).

Moreover, since the neural network models are typically over-parameterized and contain a high level of redundancy, we observe that dropping a large portion of gradients does not hamper the training. This means the length of $I_s$ can be significantly smaller than the length of $I$. This is useful in reducing the \textit{effective} sensitivity of the sum of the gradients.

We then update the parameter values for those selected using DP-SGD:
\begin{align}
    \vtheta^{(t+1)}[I_s] &=\hat\vtheta^{(t)}[I_s] - \frac{\eta_t}{B} \left[\sum_{b \in \mathcal{B}^{(t)}}\nabla^{C}[I_s, b] + \sigma C \zeta \right],
\end{align} where $\zeta \sim \Nrm(0, I)$.
For the parameters corresponding to the non-selected gradients indicated by $I_{ns}$, we simply do not update the values for them, by   
\begin{align}
    \vtheta^{(t+1)}[I_{ns}] &=\vtheta^{(t)}[I_{ns}].
\end{align} We finally update the parameter values to the new values denoted by $\vtheta^{(t+1)}$. We repeat these steps until convergence. 
%
This algorithm called, \textit{differentially private \textbf{sparse} stochastic gradient descent} (DP-SSGD) 
is given in \algoref{dp_grad_drop_var1}. Note that the sub-algorithms,  Grad-Drop and Pre-Prune are in Appendix.  
In the Experiments section we explore the two proposed tools, pre-pruning and gradient-dropping, and trade-offs between them in the form of DP-SSGD.

\input{alg.tex}


\section{Related Work}\label{sec:related_work}
Recently, there has been a flurry of efforts  on improving the performance of DP-SGD in training large-scale neural network models. These efforts are made from a wide range of angles, as described next. 
For instance, 
\cite{lyu2021dpsignsgd} developed a sign-based DP-SGD approach called DP-SignSGD for more efficient and robust training.
\cite{Papernot_Thakurta_Song_Chien_Erlingsson_2021} replaced the ReLU activations with the tempered sigmoid activation to reduce the bias caused by gradient clipping in DP-SGD and improved its performance. 
%
%
Furthermore, 
\cite{shamsabadi2022losing}
suggested modifying the loss function to promote a fast convergence in DP-SGD. 
\cite{DPNAS} developed a paradigm for 
neural network architecture search with differential privacy constraints, while
\cite{wang2019differentially} a gradient hard thresholding framework that provides good utility guarantees. 
%
%
\cite{10.1145/3469877.3490594} proposed a grouped gradient clipping mechanism to modulate the gradient weights.
\cite{scale_norm} suggested a simple architectural modification termed ScaleNorm by which an additional normalization layer is added to further improve the performance of DP-SGD. 

\cite{de2022unlocking} proposed simple techniques like weight standardization in convolutional layers, leveraging data augmentation and parameter averaging, which significantly improve the performance of DP-SGD. \cite{de2022unlocking} also shows that non-private pre-training on public data, followed by fine-tuning on private datasets yields outstanding performance. 

The transfer learning idea we use in our work also has been a popular way to save the privacy budget in DP classification. 
For instance, 
 \cite{tramer2021differentially} suggested transferring features learned from public data to classify private data with a minimal privacy cost. \cite{CVPR_Sparse} developed an architecture that consists of an encoder and a joint classifier, where these are pre-trained with public data and then fine-tuned with private data. When fine-tuning these models, \cite{CVPR_Sparse} also used the idea of neural network pruning to sparsify the gradients to update using DP-SGD.    
Our method is more general than that in  \cite{CVPR_Sparse} in that we sparsify the gradients given any architecture, while \cite{CVPR_Sparse} focuses on the particular architecture they introduced and specific fine-tuning techniques for the architecture.



Many attempts which take advantage of public data have improved the baseline performance. A recent work improved the accuracy of a privately trained classifier (ResNet-20) for CIFAR10 from $37\%$ to $60\%$ accuracy at $\epsilon=2$ \cite{Da21}.
Existing works for reducing the dimension of gradients in DP-SGD also use public data for estimating the subspaces with which the dimension of the gradients are reduced \cite{kairouz2021fast, SGD_with_Gradient_space_identification, Da21, pmlr-v139-yu21f}. \cite{random_select} suggested a random selection of gradients to reduce the dimensionality of the gradients to privatize in DP-SGD. 
Unlike these works, we also reduce the dimension of the gradients by selecting the gradients on the magnitude of each weight or weight, inspired by neural network pruning. While \cite{random_select} proposed to freeze the parameters progressively, we combine two tools, pre-pruning and gradient-dropping to improve the performance.




\section{Experiments}

The experiments are performed on three network architectures, WideResnet 16-4, WideResnet 28-10, and Resnet50, and three datasets CIFAR-10, CIFAR-100 and Imagenet. 


\subsection{Ablation study: Pre-pruning}

In the first experiment, we compare three pre-pruning methods presented in Sec.3, random, DP-SNIP, and Synflow. The three methods represent three different paradigms in terms of differentially private pre-pruning. DP-SNIP requires privatization of the pre-pruning procedure, which Synflow looks at the information flow within network but does not involve the training data. Similarly, random does not look at any data structures but arbitrarily selects the structure of the pre-pruned network, thereby does not require privatization, either. 

The results can be seen in Fig. 1. While random and Snip-DP-based pre-pruning performs better for smaller pruning rates, Synflow outperforms other benchmarks when the rates are higher. Moreover, with the increasing pruning rate, Synflow pre-pruning function has a positive correlation and overall the highest accuracy among the tested methods which is the most desirable for the DP-training.   


\begin{figure}[!tbp]
  \centering
  \begin{minipage}[b]{0.46\textwidth}
    \includegraphics[width=\textwidth]{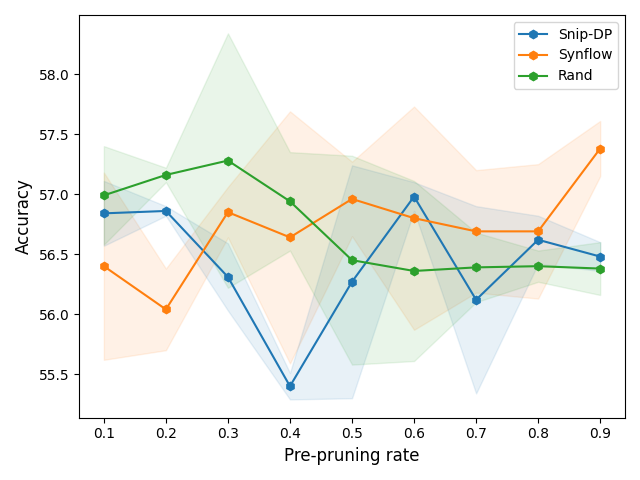}
    \caption{Comparison of pre-pruning methods on WideResnet-16-4 trained on CIFAR-10}
    \label{fig:abl_pre}
  \end{minipage}
  \hfill
  \begin{minipage}[b]{0.46\textwidth}
    \includegraphics[width=\textwidth]{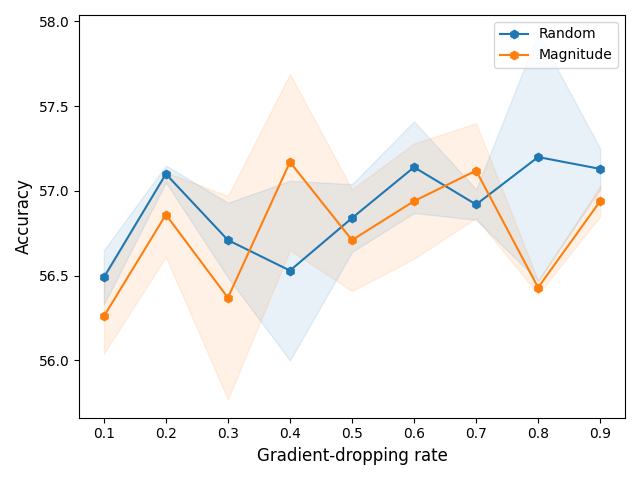}
    \caption{Comparison of grad-dropping methods on WideResnet-16-4 trained on CIFAR-10.}
    \label{fig:abl_grad}
  \end{minipage}
\end{figure}

\subsection{Ablation study: Gradient-dropping during DP-SGD}

In the second ablation study, we compare two modes of selection of gradients at every training iteration, random selection and selection based on the magnitude of the parameters. The study is performed on WideResnet-16-4 and CIFAR-10 datasets. 
The varying parameter is the pruning rate $p \in [0, 1)$ (notice we cannot remove all the parameters within a layer to prevent the layer's collapse) which specifies the fraction of parameters dropped at every layer. 

In the case of random selection, given the pruning rate $\pi_{gd}$ we select the binary mask for each layer where the set of $\pi_{gd}*|I|$ indices is selected uniformly at random. In the case of magnitude pruning, in every iteration, we sort the parameter values of every layer and drop the gradients computed with respect to the smallest parameter values. 

In this ablation study, we only study dropping the gradients during training and the network is not pre-pruned and retains the original size. The results can be seen in \figref{abl_grad}. For both random and magnitude modes, the figure shows that with the increasing pruning rate, the accuracy is also rising. Both methods perform similarly with a slight edge towards random selection.

\subsection{Combining into DP-SSGD}

Finally, to benefit from both findings, we combine two methods, pre-pruning and grad-dropping into a complete DP-SSGD method. 
In the first step, we pre-prune the network to obtain a smaller version of the original model. Subsequently, we perform training on the pruned network. In the second step, some of the gradients of the pruned network are dropped during training. As a result of the step procedure, we attempt to optimize the space of updated parameters in a way that takes advantage of both methods, one could summarize, the best of two worlds. 

Fig. 3 summarizes the results on a three-dimensional plane where we vary both the pre-pruning rate and grad dropping rate. Notice that the axis where the pre-pruning rate is 0 and the grad-dropping rate is 0 describes two particular cases where only grad-dropping or only pre-pruning is applied (the cases described in two ablation subsections). The results show a form trade-off between pre-pruning and gradient-dropping in form of the diagonal (indicated by red rises) which let us conclude that it is preferably to combine small pre-pruning rate with high-gradient-dropping rate or vice versa. Moreover, both high pre-pruning and gradient-dropping rates are preferable to both low rates. Fig. 3(top) shows in two dimensions the benefits of combining the two two tools. In the case of pre-pruning and gradient-dropping we show the base rates, in the case of combined version the base rate corresponds to the pre-prune rate which is combined with best performing gradient-dropping rate. Let us now that the combined version obtains overall the highest accuracy given all the benchmarks. These results are summaried in the next subsection where we compare the best results with each other and those existing in the literature.

\begin{figure}[!tbp]
  \label{fig:ablation}
  \centering
  \begin{minipage}[b]{0.67\linewidth}
    \includegraphics[width=\textwidth,trim={4.5cm 2cm 3cm 4.3cm},clip]{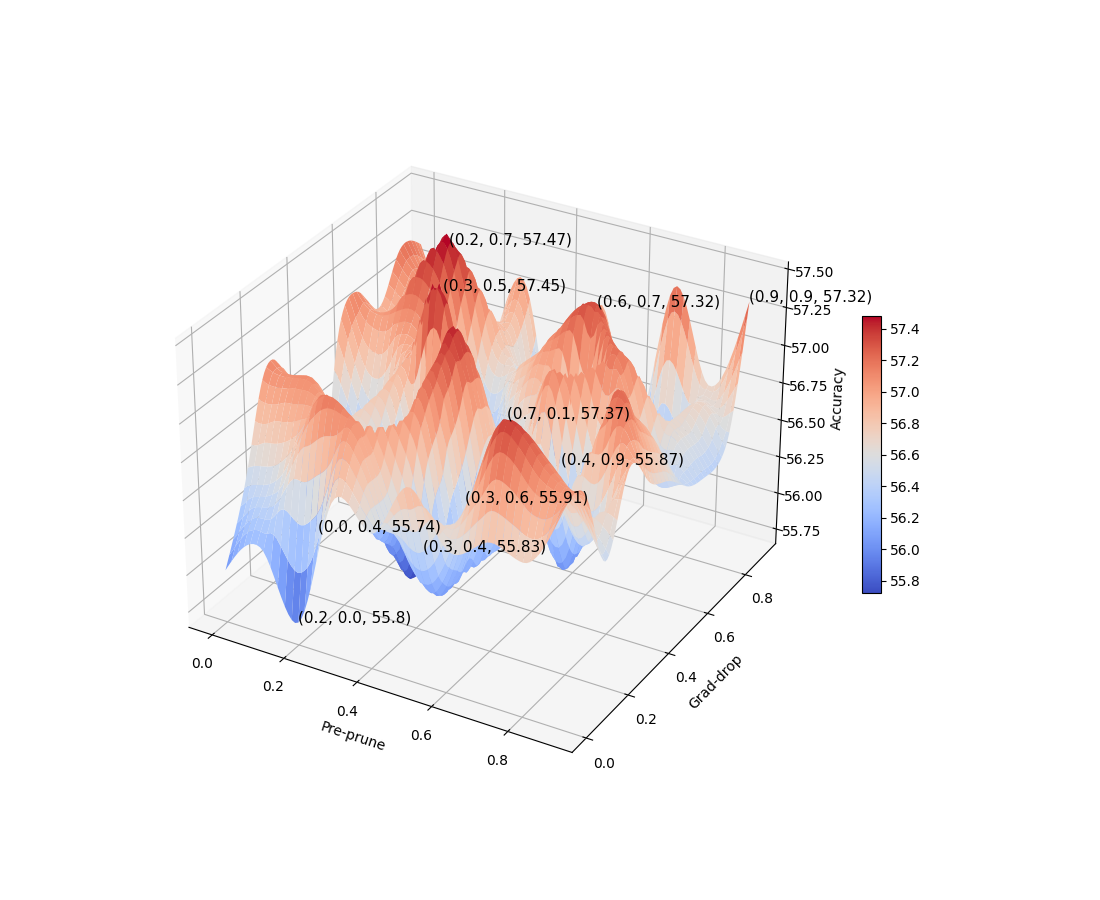}
        \vspace{-1cm}
    \label{fig:abl_pre}
  \end{minipage}
  \hfill
  \begin{minipage}[b]{0.32\linewidth}
  \vspace{-10cm}
    \includegraphics[width=\textwidth,trim={2.5cm 0 3cm 0},clip]{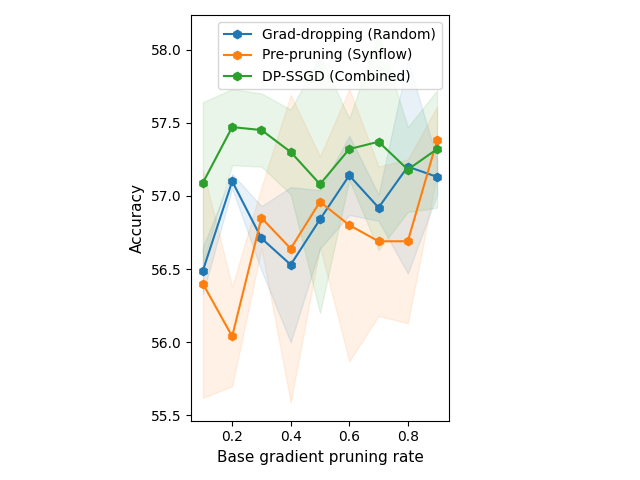}
    \caption{(left) The effect of varying pre-pruning rate and gradient-dropping rate on accuracy. (top) \hspace{3pt} Combing pre-pruning and grad-drop versus each of them in isolation.}
    \label{fig:abl_grad}
  \end{minipage}
\end{figure}

\section{Comparison}

In the case of training from scratch, we compare the benchmarks on two datasets, CIFAR-10 and Imagenet on the WideResnet 16-4 and Resnet-50. We also perform experiments on networks pre-trained on Imagenet (which is considered to be public data). In this case, we use WideResnet 28-10, which is subsequently trained CIFAR-10 and CIFAR-100 (here considered to be private data).

The results can be seen in Table 1 (no public data) and Table 2 (public data). We use \cite{de2022unlocking} as the baseline and implement the tools presented in this paper. We may see that both pre-pruning and grad-dropping on their own improve the results but when jointly used the improvements are the highest. We report the mean accuracies over five runs. Let us note that the higher levels of privacy (when the epsilon is smaller) the benefits of sparsifying the gradient are more visible.

\begin{table}[h]
    \centering
\begin{tabular}{p{70pt}ccccccccccc}
\toprule
& \multicolumn{3}{c}{CIFAR-10} && \multicolumn{2}{c}{Imagenet} && CIFAR-10 && CIFAR-100 \\
\cmidrule(lr){2-4}\cmidrule(lr){6-7}
& $\epsilon=1$ & $\epsilon=2$ & $\epsilon=4$ && $\epsilon=0.5$ & $\epsilon=1$ && $\epsilon=1$ && $\epsilon=1$ \\
\midrule
\textbf{DP-SSGD} & \textbf{57.45} & \textbf{64.03} & \textbf{68.89} && \textbf{2.64} & \textbf{6.78} && \textbf{94.84} && \textbf{67.74}\\
Pre-pruning & 57.36 & 63.67 & 68.62 && 2.61 & 6.42 && 94.75 && 68.64\\
Grad-dropping & 57.20 & 63.43 & 68.7 && 2.55 & 6.21 && 94.69 && 65.68\\
\cite{de2022unlocking} & 56.8 & 63.49 & 68.58 &&  2.43 & 6.09 && 94.65 && 67.40\\
\bottomrule
\end{tabular}
    \caption{Left two columns: CIFAR-10 / WideResnet 16-4 and Imagenet / Resnet-50 trained from scratch. Right two columns: CIFAR-10 and CIFAR-100 / WideResnet 28-10 pre-trained on Imagenet. The top-1 accuracy mean over five independent runs. For standard deviations please refer to Appendix}.
    \label{tab:my_label}
\end{table}

\section{Conclusion}

As achieving high accuracies for large neural network trained in a differentially private manner remains difficult, we propose gradient-sparsifying algorithm, DP-SSGD to facilitate DP-training. The algorithm consists of two tools, pre-pruning and gradient-dropping, which can be applied each on their own or in combination. As a result, DP-SSGD makes the DP-training more efficient, and, due to the shown improvements, it casts lights on redundancies in training neural network in DP-fashion. 

\clearpage

\section{Societal impact}
Differential privacy aims to build models by ensuring the privacy of the data it was trained on. The inherent trade-off between privacy and accuracy is difficult to overcome but DP-SSGD aims to facilitate this arduous task and produce models which are both accurate and private. That being said, a common ($\delta, \epsilon$) definition that we utiize in this paper allows for a small controlled privacy leakage. Moreover, we also partially rely on public data which not always may be available. 
{
\small
\clearpage

\setcounter {section} {0}
\renewcommand{\thesection}{\Alph{section}}

\section{Pre-prune and Grad-drop algorithms}

\input{alg_pre_prune}

\input{alg_drop}

\section{Code}

Please find the anonymized repository under this link,
\href{https://anonymous.4open.science/r/dp-ssgd1/README.md}{https://anonymous.4open.science/r/dp-ssgd1/}


\section{Further details on SNIP}

To elaborate on the paragraph `Pre-pruning with SNIP` in Sec. 3.1 we should note that the connection sensitivity $c_j$ should be a binary value where the binary value 1 stands for connection j being active. Then to take derivative, this constraint is allow to be relaxed so that $c_j$ can take any real value between (0,1). The derivative is an approximation of the effect in loss of removing connection $j$. 
Once the sensitivity is computed, only the top-k connections are retained, where k denotes the desired number of non-zero weights. The remaining connections form the pruned network. By removing the least sensitive connections, SNIP retains the connections that have a higher impact on the loss function, leading to a pruned network with minimal loss in accuracy.
However, unlike random and SynFlow pre-pruning that do not necessitate data access, SNIP pruning requires a mini-batch of data for computing the connection sensitivity.


\clearpage

\section{Ablation study of pre-pruning masks on (non-private) SGD training}

\begin{figure}[h]
    \centering
    \includegraphics[scale=0.3]{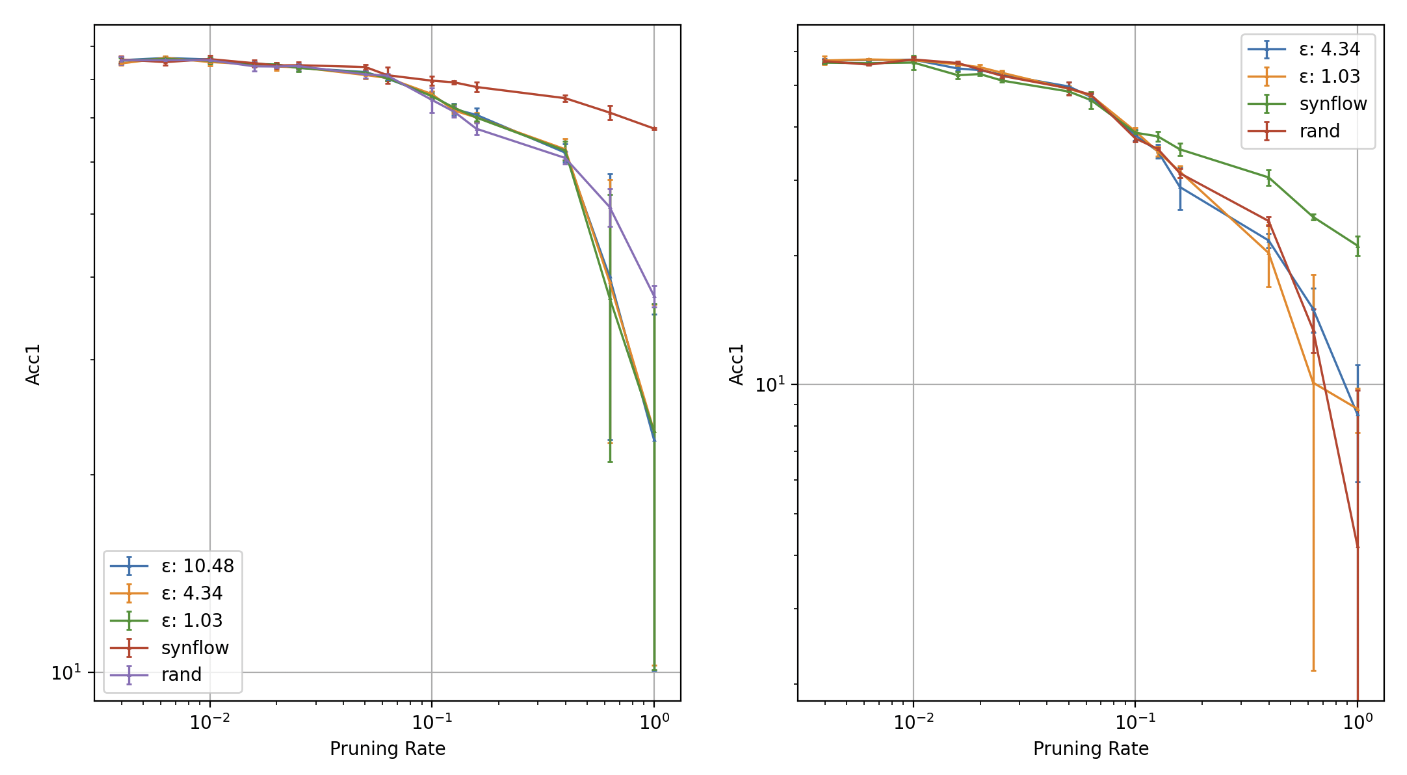}
    \caption{Top-1 accuracies with standard deviation across 5 runs of WRN16-4 on CIFAR 10 (left), CIFAR 100 (right) training from
scratch with non-dp SGD and various pruning methods. $\epsilon$ stands for DP-SNIP pruning with ($\epsilon$, $\delta=10e-5$) privacy guarantee. (\textit{GPU: RTX A4000, Total runtime: $\approx$24h for each dataset})}
    \label{fig:my_label}
\end{figure}

We can see that the epsilon does not affect the quality of the mask that much. Therefore, we select a smaller epsilon so that we do not much of our privacy budget for mask selection, and instead use it for DP training.

SynFlow pruning demonstrates superior performance across the majority of pruning rates.As the pruning rate increases, the performance disparity between SynFlow and the other two methods becomes more pronounced. The accuracy of DP-SNIP and random pruning experience a significant decline, whereas SynFlow maintains superior accuracy levels. 



\section{Standard deviations for the tables in the main text}

Here we present the same tables as in Table 1 in the Sec. 6. The tables presented in the main paper include the mean accuracy over five runs. Here, we also include the information about standard deviations. 

\begin{table}[H]
    \centering
\begin{tabular}{p{80pt}ccccccp{5pt}cccc}
\toprule
& \multicolumn{6}{c}{CIFAR-10}  \\
\cmidrule(lr){2-7}
& $\epsilon=1$ && $\epsilon=2$ && $\epsilon=4$ &\\
\midrule
\textbf{DP-SSGD} & \textbf{57.45} & $(0.23)$ & \textbf{64.03} & $(0.19)$ & \textbf{68.89} & $(0.27)$ \\
Pre-pruning & 57.36 & $(0.49)$ & 63.67 & $(0.17)$ & 68.62 & $(0.25)$  \\
Grad-dropping & 57.20 & $(0.34)$ & 63.43 & $(0.15)$ & 68.7 & $(0.31)$ \\
\cite{de2022unlocking} & 56.8 & $(0.21)$ & 63.49 & $(0.11)$ & 68.58 & $(0.22)$ \\
\bottomrule
\end{tabular}
    \caption{CIFAR-10 / WideResnet 16-4 trained from scratch, GPU time: 13 days per one run for $\epsilon=4$ (which includes the collection of results for lower $\epsilon$)}
    \label{tab:my_label}
\end{table}

\begin{table}[H]
    \centering
\begin{tabular}{p{80pt}cccc}
\toprule

& \multicolumn{4}{c}{Imagenet} \\

\cmidrule(lr){2-5} 
 & $\epsilon=0.5$ && $\epsilon=1$ \\
\midrule
\textbf{DP-SSGD}  & \textbf{2.64} & $(0.22)$ & \textbf{6.78}&$(0.18)$ \\
Pre-pruning  & 2.61 & $(0.15)$ & 6.42&$(0.17)$ \\
Grad-dropping  & 2.55 & $(0.14)$ & 6.21& $(0.19)$ \\
\cite{de2022unlocking} &  2.43 & $(0.17)$ & 6.09&$(0.11)$ \\
\bottomrule
\end{tabular}
    \caption{Imagenet / Resnet-50 trained from scratch, GPU time: 10 days per one run for $\epsilon=1$}
    \label{tab:my_label}
\end{table}
\vspace{-1cm}

\begin{table}[H]
    \centering
\begin{tabular}{p{80pt}ccp{5pt}cc}
\toprule
& \multicolumn{2}{c}{CIFAR-10} && \multicolumn{2}{c}{CIFAR-100} \\
\cmidrule(lr){2-3}\cmidrule(lr){5-6} 
 & $\epsilon=1$ &  && $\epsilon=1$ \\
\midrule
DP-SSGD  & 94.84 & 
{\color{black}(0.16)} & & 67.74 & 
{\color{black}(0.31)}\\
Pre-pruning  & 94.75 & {\color{black}(0.14)} & & 68.64 &
{\color{black}(0.28)}\\
Grad-dropping & 94.69 & {\color{black}(0.04)} & & 65.68 & {\color{black}(0.64)} \\
\cite{de2022unlocking}  & 94.65& 
{\color{black}(0.03)} & &  67.40 &
{\color{black}(0.22)} \\
\bottomrule
\end{tabular}
    \caption{CIFAR-10 / WRN-28-10 pre-trained on Imagenet, GPU time: 4 days per one run for $\epsilon=1$}
    \label{tab:my_label}
\end{table}
\vspace{-1cm}

\section{Detailed results for various pre-pruning and grad-dropping rates of the DP-SGD training}

In this section, we present detailed numerical results presented visually in Fig. 3 in the main text. The training was performed from scratch with CIFAR-10 on WideResnet 16-4. The details include both accuracy and standard deviation.

The numbers in bold are the highest accuracies for given pre-pruing level selected among different grad-dropping pruning rates. They correcpond to red peaks in Fig. 3. Correspondingly, the numbers in italics are the lowest accuracies for given pre-pruning rates and they are represented by blue troughs in Fig. 3. Please also refer to Sec. 5.3 in the main text.

\begin{tabular}{cccccccccccc}
\toprule
pre-prune $\downarrow$ \\ grad-drop $\rightarrow$ $$& 0.0 &0.1 & 0.2 & 0.3 & 0.4 & 0.5 & 0.6 & 0.7 & 0.8 & 0.9 \\ \midrule

0.00 & 56.00 & 56.49 & 57.10 & 56.71 & \textit{55.74} & 56.84 & 57.14 & 56.92 & \textbf{57.20} & 57.13 & \\ 
 & (0.55) & (0.16) & (0.05) & (0.22) & (0.63) & (0.37) & (0.27) & (0.09) & (0.73) & (0.12) & \\ \midrule

0.10 & \textit{56.23} & 56.62 & \textbf{57.09} & 56.74 & 56.38 & 56.36 & 57.06 & 56.93 & 57.02 & 56.57 & \\ 
 & (0.68) & (0.22) & (0.55) & (0.43) & (0.16) & (0.59) & (0.07) & (0.13) & (0.24) & (0.33) & \\ \midrule
 
0.20 & \textit{55.8} & 57.1 & 56.8 & 56.52 & 56.74 & 56.96 & 56.96 & \textbf{57.47} & 56.89 & 56.56 & \\
 & (0.35) & (0.77) & (0.33) & (0.71) & (0.2) & (0.31) & (0.89) & (0.26) & (0.08) & (0.75) & \\ \midrule
 
0.30 & 56.67 & 57.09 & 56.24 & 55.94 & \textit{55.83} & \textbf{57.45} & 55.91 & 56.5 & 56.8 & 57.03 & \\
 & (0.36) & (0.24) & (0.45) & (0.14) & (0.47) & (0.09) & (0.34) & (0.70) & (0.31) & (0.24) & \\ \midrule
 
0.40 & 56.55 & 56.53 & 57.08 & 57.3 & \textbf{57.27} & 56.28 & 56.48 & 56.8 & 56.2 & \textit{55.87} & \\
 & (0.93) & (0.54) & (0.31) & (0.29) & (0.33) & (0.38) & (0.22) & (0.49) & (0.14) & (0.72) & \\ \midrule
 
0.50 & 56.92 & 56.12 & \textit{56.11} & 56.37 & 56.68 & 56.51 & 56.89 & \textbf{57.08} & 56.52 & 57.04 & \\
 & (0.29) & (0.12) & (0.25) & (1.15) & (0.62) & (0.2) & (0.54) & (0.88) & (0.29) & (0.01) & \\ \midrule
 
0.60 & 56.6 & 56.73 & 56.66 & \textit{56.28} & 56.42 & 57.14 & 56.34 & \textbf{57.32} & 56.76 & 56.32 & \\
 & (0.8) & (0.44) & (0.25) & (0.62) & (0.4) & (0.52) & (0.48) & (0.21) & (0.38) & (0.37) & \\ \midrule
 
0.70 & 56.71 & \textbf{57.37} & 56.73 & 56.62 & \textit{56.05} & 56.34 & 57.19 & 56.76 & 56.17 & 57.28 & \\
 & (0.46) & (0.74) & (0.27) & (0.35) & (0.33) & (0.51) & (0.79) & (0.49) & (0.57) & (0.24) & \\ \midrule
 
0.80 & 56.66 & 57.15 & 56.8 & \textit{56.41} & \textbf{57.18} & 56.94 & 56.48 & 56.69 & 56.68 & 56.7 & \\
 & (0.52) & (0.41) & (1.2) & (0.4) & (0.29) & (0.27) & (0.05) & (0.8) & (0.74) & (0.32) & \\ \midrule
 
0.90 & 57.06 & 57.04 & 56.48 & 56.92 & 56.75 & 56.62 & 56.46 & \textit{56.4} & 56.47 & \textbf{57.32} & \\
 & (0.49) & (0.12) & (0.59) & (0.07) & (0.43) & (0.3) & (0.32) & (0.19) & (0.41) & (0.4) & \\ \midrule
\bottomrule

\end{tabular}

\section{Detailed results for various pruning rates of the DP-SGD training of the pre-trained network }

Here, we present detailed results for varying pre-pruning and grad-dropping rates for WideResnet 28-10 pre-trained with Imagenet and trained on CIFAR-10 and CIFAR-100. The best results are presented in Table 1 in the main text. Here we do not combine the two. The results are shown in Table ~\ref{tab: WRN2810}. 

\begin{table}[H]
\centering
    \begin{tabular}{c c cc cc}
      \hline
      pruning methods & Sparsity & CIFAR10 && CIFAR100&\\
      \hline
      base (no pruning, \\ \cite{de2022unlocking}) & 0.0 & 94.65 & (0.18) & 67.4 & (0.31) \\
      \hline
      Pre-pruning & 0.2  &  94.75&(0.14) &  67.64 & (0.28)\\
      (Synflow)        & 0.4  &  94.60&(0.13)  &  67.15 &(0.40)\\
              & 0.6  &  94.69&(0.12) &  67.03 &(0.47)\\
              & 0.8  &  94.35&(0.20) &  66.48 &(0.37)\\
      \hline
      Grad-dropping    & 0.2  &  94.69&(0.04) &  65.85&(0.64)\\
      (Random)        & 0.4  &  94.30&(0.08)  &  62.84&(0.10) \\
              & 0.6  &  94.05&(0.10) &  58.08&(0.60) \\
              & 0.8  &  92.81&(0.24)  &  47.56&(1.50) \\
      \hline
    \end{tabular}
  \caption{Test results on CIFAR10 and CIFAR100 with WRN28-10 pretrained on Imagenet (down-sampled to 32 x 32). We report the median accuracy across 4 runs followed by standard deviation. Privacy budget for training are fixed at ($\epsilon_t = 1, \delta=1e-5$) (\textit{GPU: NVIDIA Tesla V100 16GB, Runtime: 3-4 days for each run})}
  \label{tab: WRN2810}
\end{table}

\clearpage
\bibliographystyle{plainnat}
\bibliography{bib}
\end{document}

%% file: notations.tex
\newcommand{\punt}[1]{}

\usepackage{amsthm}
\usepackage{xcolor}
\usepackage{hyperref}

\renewcommand{\eqref}[1]{eq.~\ref{eq:#1}}
\newcommand{\Nrm}{\mathcal{N}}

\newcommand{\figref}[1]{Fig.~\ref{fig:#1}}  
\newcommand{\secref}[1]{Sec.~\ref{sec:#1}}
\newcommand{\algoref}[1]{Algorithm~\ref{algo:#1}}  

\newcommand{\Dat}{\mathcal{D}}

\newcommand{\vone}{\mathbf{1}} 

\newcommand{\ve}{\mathbf{e}}

\newcommand{\vtheta}{\mathbf{\ensuremath{\bm{\theta}}}}
\newcommand{\LL}{\ensuremath{\mathcal{L}}}

\newcommand{\vn}{\mathbf{n}}

\newcommand{\vg}{\mathbf{g}}

\def\Pr{\ensuremath{\text{Pr}}}

%% file: alg.tex
\begin{algorithm}
\caption{DP-SSGD}\label{algo:dp_grad_drop_var1}
\begin{algorithmic}[1]
\Require Dataset $\Dat$, pruning rates $\pi_{pp}$, $\pi_{gd}$ for pre-pruning and grad-dropping, norm clipping hyperparameter $C_{pp}, C_{gd}$ for pre-pruning (DP-SNIP only) and grad-dropping, $\epsilon, \delta$ values for pre-pruning and grad-dropping, indices of weights (parameterers) $I$. 
\Ensure Differentially private model gradients (parameters) 
\State
\Function{DP-SGD}{$\sigma, C, \nabla, \mathcal{B}$}
    \State We compute the average gradient matrix: $\vg = \frac{1}{B} \sum_{b=1}^S \nabla[:,b]$, where $b \in \mathcal{B}$.
   \State  We ensure  $  \|\nabla[I_s]\|_2 \leq C$ by clipping the L2-norm by $C$.
    \State We privatize the gradients in the selected index set,  $\nabla \leftarrow \nabla[I_s] + \vn$, where $\vn \sim \Nrm(0, \sigma^2 C^2 I)$.
    
\Return {$\nabla$}
\EndFunction
\State
\Function{DP-Sparse-SGD}{}

     \State $I_{pp}, \epsilon_{pp} \leftarrow$ \Call{Pre-prune}{$\pi_{pp}, C_{pp}$}, 
    \State Calculate privacy parameter $\sigma$ from $\epsilon_{gd}$, where 
      $\epsilon_{gd} = \epsilon-\epsilon_{pp}$ if DP-SNIP and 
      $\epsilon$ otherwise
    \For{batch $\mathcal{B} \in \Dat $}        

     \State $I_{gd} \leftarrow$ \Call{Grad-drop}{$\pi_{gd}, C_{gd}$}
     \State $I_{rm} \leftarrow I_{pp} \cup I_{gd}$
    \State $I \leftarrow \nabla[I_{rm}] = 0$
    \State $\nabla_{DP-SSGD} \leftarrow$ \Call{DP-SGD}{}
    \EndFor

\Return{$\nabla_{DP-SSGD}$} 
\EndFunction

\end{algorithmic}
\end{algorithm}

%% file: alg_pre_prune.tex

\begin{algorithm}[H]
\caption{DP-Pre-Prune}\label{algo:pre_prune}
\begin{algorithmic}[1]
\Require A base model with parameter indices $I$, pre-pruning rate $\pi_{pp}$, criterion: $\leftarrow$ {Random, Synflow, DP-SNIP}, (if DP-SNIP) dataset $\Dat$, clipping norm $C_p$, privacy parameters, $\epsilon_p$ and $\delta_p$.
\Ensure Pre-pruned network \\
\Function{DP-Pre-prune}{Base model, criterion, $\pi_{pp}$, $\Dat$, $C_p$,  $\epsilon_p$,  $\delta_p$}
    \If{criterion == Random}
     \State $I_{pp}$ $\leftarrow$ Remove parameters uniformly at random with pruning rate $\pi_{pp}$
    \ElsIf{criterion == Synflow}
     \State $I_{pp}$ $\leftarrow$ Remove parameters based on Eq. 2 with pruning rate $\pi_{pp}$ 
    \ElsIf {criterion == DP-SNIP}
    \State $I_{pp}$ $\leftarrow$ DP-SNIP($\pi_{pp}$, $\Dat$, $C_p$,  $\epsilon_p$,  $\delta_p$) 
    \EndIf \\
    \Return model pre-pruned indices {$I_{pp}$}
\EndFunction
\State
\Function{DP-SNIP}{$p_p$, $\Dat$, $C_p$,  $\epsilon_p$,  $\delta_p$}
\State \textbf{Require: }Loss function $\mathcal{L}$, Connectivity indicator \textbf{c}, connection j  
\State $\theta \xleftarrow{}$ Initialization
\State $g_j(\theta;\Dat) \xleftarrow{} \left.\frac{\partial \mathcal{L}(\textbf{c} \odot \theta ; \Dat)}{\partial c_j}\right\vert_{\textbf{c}=1},\quad\forall j \in \{1, ..., m\}\hfill \triangleright \text{Effect of connection j on the loss}$
\State $\tilde{g}_j \xleftarrow{} \frac{1}{B}clip_{C_p}(g_j(w; \Dat)) + \frac{\sigma C_p}{B}\xi,\quad\forall j \in \{1, ..., m\}\hfill \triangleright B = |\Dat|$
\State $s_j \xleftarrow{} \frac{|\tilde{g}_j|}{\sum_{k=1}^m |\tilde{g}_j|},\quad \forall j \in \{1, ..., m\} \hfill \triangleright \text{Connection sensitivity}$
\State  $\hat{s} \xleftarrow{} SortDescending(s)$
\State $c_j \xleftarrow{} \mathbbm{1}[s_j - \hat{s_\kappa} \geq 0], \quad \forall j \in \{1, ..., m\} \hfill \triangleright \text{Pruning: choose the top connections}$
\State $\theta^* \xleftarrow{} argmin_{\theta\in \mathbb{R}^m} L(\textbf{c} \odot \theta; \Dat) \hfill \triangleright \text{Regular training}$
\State $\theta^* \xleftarrow{} c \odot \theta^*$

\Return{$\theta^*$}
\EndFunction

\end{algorithmic}
\end{algorithm}

    \label{fig:algo}

%% file: alg_drop.tex
\begin{figure}[H]

\begin{algorithm}[H]
\caption{Grad-Drop}\label{algo:DPGDSGD}
\begin{algorithmic}[1]
\Require (Possibly pre-pruned) model with corresponding indices $I$. gradient-dropping rate $\pi_{gd}$, criterion $\leftarrow$ {Random, Magnitude}. 
\Ensure Indices to drop (or keep) in updating \\
\State
\Function{Select-indices}{$C$, criterion, weights}
    \If{criterion == Random}
     \State $I_{ns}$ $\leftarrow \pi_{gd} \cdot len(I)$ indices of $I$, selected uniformly at random 
    \ElsIf{criterion == Magnitude}
     \State $I_{as} \leftarrow$ argsort(weights)
    \State $I_{ns} \leftarrow$ bottom $\pi_{gd} \cdot len(I)$ indices of $I_{as}$ 
    \EndIf

\State $I_s \leftarrow I \setminus I_{ns}$ \\
\Return {$I_{ns}, I_{s}$}
\EndFunction




%
\end{algorithmic}
\end{algorithm}

    \caption{Grad-drop algorithm. The indices for dropping gradients are selected at every iteration. As the weights in each training step are already DP, selecting indices based on the magnitude does not incur privacy loss.}
    \label{fig:algo}
\end{figure}

%% file: arxiv.bbl
\begin{thebibliography}{25}
\providecommand{\natexlab}[1]{#1}
\providecommand{\url}[1]{\texttt{#1}}
\expandafter\ifx\csname urlstyle\endcsname\relax
  \providecommand{\doi}[1]{doi: #1}\else
  \providecommand{\doi}{doi: \begingroup \urlstyle{rm}\Url}\fi

\bibitem[Abadi et~al.(2016)Abadi, Chu, Goodfellow, McMahan, Mironov, Talwar,
  and Zhang]{DP_SGD}
Martin Abadi, Andy Chu, Ian Goodfellow, H.~Brendan McMahan, Ilya Mironov, Kunal
  Talwar, and Li~Zhang.
\newblock Deep learning with differential privacy.
\newblock In \emph{Proceedings of the 2016 ACM SIGSAC Conference on Computer
  and Communications Security}, CCS ’16, page 308–318, New York, NY, USA,
  2016. Association for Computing Machinery.
\newblock ISBN 9781450341394.
\newblock \doi{10.1145/2976749.2978318}.

\bibitem[Aji and Heafield(2017)]{Grad_Drop}
Alham~Fikri Aji and Kenneth Heafield.
\newblock Sparse communication for distributed gradient descent.
\newblock In \emph{Proceedings of the 2017 Conference on Empirical Methods in
  Natural Language Processing}, pages 440--445, Copenhagen, Denmark, September
  2017. Association for Computational Linguistics.
\newblock \doi{10.18653/v1/D17-1045}.
\newblock URL \url{https://aclanthology.org/D17-1045}.

\bibitem[Balle et~al.(2022)Balle, Berrada, De, Hayes, Smith, and
  Stanforth]{jax-privacy2022github}
Borja Balle, Leonard Berrada, Soham De, Jamie Hayes, Samuel~L Smith, and Robert
  Stanforth.
\newblock {JAX}-{P}rivacy: Algorithms for privacy-preserving machine learning
  in jax, 2022.
\newblock URL \url{http://github.com/deepmind/jax_privacy}.

\bibitem[Cheng et~al.(2021)Cheng, Wang, Zhang, Chen, Wang, and Cheng]{DPNAS}
Anda Cheng, Jiaxing Wang, Xi~Sheryl Zhang, Qiang Chen, Peisong Wang, and Jian
  Cheng.
\newblock {DPNAS:} neural architecture search for deep learning with
  differential privacy.
\newblock \emph{CoRR}, abs/2110.08557, 2021.
\newblock URL \url{https://arxiv.org/abs/2110.08557}.

\bibitem[De et~al.(2022)De, Berrada, Hayes, Smith, and Balle]{de2022unlocking}
Soham De, Leonard Berrada, Jamie Hayes, Samuel~L Smith, and Borja Balle.
\newblock Unlocking high-accuracy differentially private image classification
  through scale.
\newblock \emph{arXiv preprint arXiv:2204.13650}, 2022.

\bibitem[Dwork and Roth(2014)]{Dwork14}
Cynthia Dwork and Aaron Roth.
\newblock The algorithmic foundations of differential privacy.
\newblock \emph{Found. Trends Theor. Comput. Sci.}, 9:\penalty0 211--407,
  August 2014.
\newblock ISSN 1551-305X.
\newblock \doi{10.1561/0400000042}.
\newblock URL \url{http://dx.doi.org/10.1561/0400000042}.

\bibitem[Dwork et~al.(2006)Dwork, Kenthapadi, McSherry, Mironov, and
  Naor]{dwork2006our}
Cynthia Dwork, Krishnaram Kenthapadi, Frank McSherry, Ilya Mironov, and Moni
  Naor.
\newblock {Our data, ourselves: Privacy via distributed noise generation}.
\newblock In \emph{Annual International Conference on the Theory and
  Applications of Cryptographic Techniques}, pages 486--503. Springer, 2006.

\bibitem[Frankle and Carbin(2018)]{frankle2018lottery}
Jonathan Frankle and Michael Carbin.
\newblock The lottery ticket hypothesis: Finding sparse, trainable neural
  networks.
\newblock \emph{arXiv preprint arXiv:1803.03635}, 2018.

\bibitem[Kairouz et~al.(2021)Kairouz, Ribero, Rush, and
  Thakurta]{kairouz2021fast}
Peter Kairouz, Mónica Ribero, Keith Rush, and Abhradeep Thakurta.
\newblock Fast dimension independent private adagrad on publicly estimated
  subspaces, 2021.

\bibitem[Klause et~al.(2022)Klause, Ziller, Rueckert, Hammernik, and
  Kaissis]{scale_norm}
Helena Klause, Alexander Ziller, Daniel Rueckert, Kerstin Hammernik, and
  Georgios Kaissis.
\newblock Differentially private training of residual networks with scale
  normalisation, 2022.
\newblock URL \url{https://arxiv.org/abs/2203.00324}.

\bibitem[Kurakin et~al.(2022)Kurakin, Song, Chien, Geambasu, Terzis, and
  Thakurta]{jax_dp}
Alexey Kurakin, Shuang Song, Steve Chien, Roxana Geambasu, Andreas Terzis, and
  Abhradeep Thakurta.
\newblock Toward training at imagenet scale with differential privacy, 2022.
\newblock URL \url{https://arxiv.org/abs/2201.12328}.

\bibitem[Lee et~al.(2019)Lee, Gresele, Park, and Muandet]{Lee_et_al_2019}
Si~Kai Lee, Luigi Gresele, Mijung Park, and Krikamol Muandet.
\newblock Private causal inference using propensity scores.
\newblock \emph{CoRR}, abs/1905.12592, 2019.
\newblock URL \url{http://arxiv.org/abs/1905.12592}.

\bibitem[Liu et~al.(2021)Liu, Li, Liu, Wang, Ge, and
  Wang]{10.1145/3469877.3490594}
Haolin Liu, Chenyu Li, Bochao Liu, Pengju Wang, Shiming Ge, and Weiping Wang.
\newblock Differentially private learning with grouped gradient clipping.
\newblock In \emph{ACM Multimedia Asia}, MMAsia '21, New York, NY, USA, 2021.
  Association for Computing Machinery.
\newblock ISBN 9781450386074.
\newblock \doi{10.1145/3469877.3490594}.
\newblock URL \url{https://doi.org/10.1145/3469877.3490594}.

\bibitem[Luo et~al.(2021)Luo, Wu, Adeli, and Fei-Fei]{CVPR_Sparse}
Zelun Luo, Daniel~J. Wu, Ehsan Adeli, and Li~Fei-Fei.
\newblock Scalable differential privacy with sparse network finetuning.
\newblock In \emph{2021 IEEE/CVF Conference on Computer Vision and Pattern
  Recognition (CVPR)}, pages 5057--5066, 2021.
\newblock \doi{10.1109/CVPR46437.2021.00502}.

\bibitem[Lyu(2021)]{lyu2021dpsignsgd}
Lingjuan Lyu.
\newblock Dp-signsgd: When efficiency meets privacy and robustness, 2021.

\bibitem[Papernot et~al.(2021)Papernot, Thakurta, Song, Chien, and
  Erlingsson]{Papernot_Thakurta_Song_Chien_Erlingsson_2021}
Nicolas Papernot, Abhradeep Thakurta, Shuang Song, Steve Chien, and Úlfar
  Erlingsson.
\newblock Tempered sigmoid activations for deep learning with differential
  privacy.
\newblock \emph{Proceedings of the AAAI Conference on Artificial Intelligence},
  35\penalty0 (10):\penalty0 9312--9321, May 2021.
\newblock URL \url{https://ojs.aaai.org/index.php/AAAI/article/view/17123}.

\bibitem[Shamsabadi and Papernot(2022)]{shamsabadi2022losing}
Ali~Shahin Shamsabadi and Nicolas Papernot.
\newblock Losing less: A loss for differentially private deep learning, 2022.
\newblock URL \url{https://openreview.net/forum?id=u7PVCewFya}.

\bibitem[Tanaka et~al.(2020)Tanaka, Kunin, Yamins, and Ganguli]{Synflow}
Hidenori Tanaka, Daniel Kunin, Daniel L.~K. Yamins, and Surya Ganguli.
\newblock Pruning neural networks without any data by iteratively conserving
  synaptic flow.
\newblock In Hugo Larochelle, Marc'Aurelio Ranzato, Raia Hadsell,
  Maria{-}Florina Balcan, and Hsuan{-}Tien Lin, editors, \emph{Advances in
  Neural Information Processing Systems 33: Annual Conference on Neural
  Information Processing Systems 2020, NeurIPS 2020, December 6-12, 2020,
  virtual}, 2020.
\newblock URL
  \url{https://proceedings.neurips.cc/paper/2020/hash/46a4378f835dc8040c8057beb6a2da52-Abstract.html}.

\bibitem[Tramer and Boneh(2021)]{tramer2021differentially}
Florian Tramer and Dan Boneh.
\newblock Differentially private learning needs better features (or much more
  data).
\newblock In \emph{International Conference on Learning Representations}, 2021.
\newblock URL \url{https://openreview.net/forum?id=YTWGvpFOQD-}.

\bibitem[Wang and Gu(2019)]{wang2019differentially}
Lingxiao Wang and Quanquan Gu.
\newblock Differentially private iterative gradient hard thresholding for
  sparse learning.
\newblock In \emph{28th International Joint Conference on Artificial
  Intelligence}, 2019.

\bibitem[Wang et~al.(2019)Wang, Balle, and Kasiviswanathan]{wang2019subsampled}
Yu-Xiang Wang, Borja Balle, and Shiva~Prasad Kasiviswanathan.
\newblock Subsampled r{\'e}nyi differential privacy and analytical moments
  accountant.
\newblock PMLR, 2019.

\bibitem[Yu et~al.(2021{\natexlab{a}})Yu, Zhang, Chen, and Liu]{Da21}
Da~Yu, Huishuai Zhang, Wei Chen, and Tie-Yan Liu.
\newblock Do not let privacy overbill utility: Gradient embedding perturbation
  for private learning.
\newblock In \emph{ICLR}, 2021{\natexlab{a}}.
\newblock URL \url{https://openreview.net/forum?id=7aogOj_VYO0}.

\bibitem[Yu et~al.(2021{\natexlab{b}})Yu, Zhang, Chen, Yin, and
  Liu]{pmlr-v139-yu21f}
Da~Yu, Huishuai Zhang, Wei Chen, Jian Yin, and Tie-Yan Liu.
\newblock Large scale private learning via low-rank reparametrization.
\newblock In Marina Meila and Tong Zhang, editors, \emph{Proceedings of the
  38th International Conference on Machine Learning}, volume 139 of
  \emph{Proceedings of Machine Learning Research}, pages 12208--12218. PMLR,
  18--24 Jul 2021{\natexlab{b}}.
\newblock URL \url{https://proceedings.mlr.press/v139/yu21f.html}.

\bibitem[Zhou et~al.(2020)Zhou, Wu, and
  Banerjee]{SGD_with_Gradient_space_identification}
Yingxue Zhou, Zhiwei~Steven Wu, and Arindam Banerjee.
\newblock Bypassing the ambient dimension: Private {SGD} with gradient subspace
  identification.
\newblock \emph{CoRR}, abs/2007.03813, 2020.
\newblock URL \url{https://arxiv.org/abs/2007.03813}.

\bibitem[Zhu and Blaschko(2021)]{random_select}
Junyi Zhu and Matthew~B. Blaschko.
\newblock Differentially private {SGD} with sparse gradients.
\newblock \emph{CoRR}, abs/2112.00845, 2021.
\newblock URL \url{https://arxiv.org/abs/2112.00845}.

\end{thebibliography}
